\documentclass[runningheads]{llncs}

\usepackage{cite}
\usepackage{amsmath,amssymb,amsfonts}
\usepackage{algorithmic}
\usepackage{graphicx}
\usepackage{textcomp}
\usepackage{booktabs} 
\usepackage{array}    
\usepackage{multirow} 
\usepackage{makecell}
\usepackage{amsmath}
\usepackage{algorithm,algorithmic}
\usepackage{hyperref}
\hypersetup{hidelinks=true}
\usepackage{textcomp}

\usepackage[T1]{fontenc}
\usepackage{graphicx}
\begin{document}
\title{GEPD: GAN-Enhanced Generalizable Model for EEG-Based Detection of Parkinson's Disease}
%
%
\author{Qian Zhang\inst{1}, Ruilin Zhang\inst{1}, Biaokai Zhu\inst{2}, Xun Han\inst{3}, Jun Xiao\inst{1}, Yifan Liu\inst{1}, Zhe Wang\inst{1}}
\authorrunning{Zhang et al.}
%
\institute{Key Laboratory of Smart Manufacturing in Energy Chemical Process, Ministry of Education,East China University of Science and Technology, Shanghai, China \and
Shanxi Police College, Taiyuan, Shanxi, China \and Intelligent Policing Key Laboratory of Sichuan Province, Sichuan Police College, Luzhou, Sichuan, China}

\maketitle              
\begin{abstract}
	Electroencephalography has been established as an effective method for detecting Parkinson's disease, typically diagnosed early.
	Current Parkinson's disease detection methods have shown significant success within individual datasets, however, the variability in detection methods across different EEG datasets and the small size of each dataset pose challenges for training a generalizable model for cross-dataset scenarios. 
	To address these issues, this paper proposes a GAN-enhanced generalizable model, named GEPD, specifically for EEG-based cross-dataset classification of Parkinson's disease.
	First, we design a generative network that creates fusion EEG data by controlling the distribution similarity between generated data and real data.
	In addition, an EEG signal quality assessment model is designed to ensure the quality of generated data great.
	Second, we design a classification network that utilizes a combination of multiple convolutional neural networks to effectively capture the time-frequency characteristics of EEG signals, while maintaining a generalizable structure and ensuring easy convergence.
	This work is dedicated to utilizing intelligent methods to study pathological manifestations, aiming to facilitate the diagnosis and monitoring of neurological diseases.
	The evaluation results demonstrate that our model performs comparably to state-of-the-art models in cross-dataset settings, achieving an accuracy of 84.3\% and an F1-score of 84.0\%, showcasing the generalizability of the proposed model.
\end{abstract}


\section{Introduction}

Electroencephalography (EEG) is a non-invasive technique for recording brain activity, widely used in neuroscience and clinical diagnostics\cite{c:5,c:6}. 
Parkinson's disease (PD) is a neurodegenerative disorder marked by dopaminergic neuron loss and Lewy body formation \cite{c:1,c:2}.
Therefore, EEG can provide promising approach for early detection and precise intervention of PD.

With the advancement of deep learning technology, many researchers utilize convolutional neural networks for PD detection.
A multi-layer 1D-CNN architecture is designed and it treats EEG signals as sequential data for disease detection \cite{c:11}.
To achieve better performance, a 2D-CNN architecture employs a smoothed pseudo-Wigner Ville distribution method to convert EEG signals into time-frequency feature maps \cite{c:12}.
However, these models are limited to single datasets, making these models difficult to apply in real-world scenarios.
Therefore, this paper aims to design a generalizable model for PD classification across different datasets. 

To achieve the above goal, two main challenges exist.
On the one hand, due to the limitations of medical data privacy protection, EEG datasets usually have a small size.
Thus, it is challenging to obtain a large EEG dataset, which cannot support training in a large-scale generative model and thus result in insufficient performance.
Although using Generative Adversarial Networks (GANs) to generate EEG data may be an effective method to alleviate this challenge \cite{c:17,c:19,c:20}, these researches have only focused on generating metrics and has not validated its effectiveness on downstream tasks.
On the other hand, how to design and train a generalizable and high-performance model for cross-dataset PD classification is challenging, 
because different datasets have variations in collection standards and subject conditions.
In addition, the performance of a model training and testing on two different datasets is lower than that of on one single dataset \cite{c:13}.
So some new methods like Meta learning \cite{c:14} are used to improve the cross-dataset detection accuracy.
However, these methods still rely on pre-training on the datasets instead of the scenario of zero-shot, which lack practical significance and broader applications. 

To overcome these challenges, this paper proposes a GAN-enhanced generalizable model, named GEPD, for EEG-based cross-dataset classification of PD.
First, it designs a generative adversarial network to construct fusion data, utilizing Jensen–Shannon divergence between the generated data and the original data. This network ensures the generation of high-quality data, which facilitates dataset expansion for subsequent classification tasks.
Second, it calculates similarity of EEG signals between channels and selects key channels in order to reduce the number of channels for further classification.
Third, based on the fusion data, it designs a network featuring parameter-efficient dilated and depthwise convolutions.
To improve cross-dataset generalizability, the network is trained with cosine annealing learning rate and multiple regularization, maintaining accuracy on single dataset while enhancing the performance in cross-dataset settings.
The extensive evaluations demonstrate that GEPD achieves comparable performance and enables effective balance between learning on one single dataset and cross-dataset scenarios.
This work is dedicated to utilizing intelligent methods to study pathological manifestations, aiming to facilitate the diagnosis and monitoring of neurological diseases.
The main contributions are listed as follows.
\begin{itemize}
	\item We propose a GAN-enhanced model GEPD, which utilizes data augmentation and generalizable classification for PD detection.
	To the best	of our knowledge, this is the first work that realizes EEG-based zero-shot cross-dataset detection of PD.
	
	\item 
	We design a channel pruning method aimed at reducing the redundancy in the expanded generated data, by adopting the mindset that less is more, to improve PD classification performance.
	
	
	\item We implement GEPD and conduct extensive experiments for evaluation. The results demonstrate that GEPD has comparable performance with an accuracy of 84.3\% and an F1-score of 84.0\%, respectively, and promising generalization ability on cross-dataset settings.
	
\end{itemize}


\section{Related Work}
This section discusses two main topics related to this work: GAN-based methods for augmenting EEG datasets and EEG-based PD classification.

Generative adversarial networks (GANs) are a typical type of network to generate new data with a distribution similar to the real data, especially in the field of image generation \cite{c:31,c:30,c:29}.
Recently, many researchers apply GANs in the field of EEG signal generation.
For exmaple, inspired by WaveGAN \cite{Donahue2018AdversarialAS}, 
R2WaveGAN \cite{c:19} combines spectral and anti-collapse regularizers to generate data.
Furthermore, researchers use the generated EEG signals for downstream tasks.
For emotion recognition, cWGAN \cite{c:16} applies mixed generated data and achieves a 2.97\% performance improvement on SEED dataset.
DEVAE-GAN \cite{c:21}, a dual-encoder variational autoencoder GAN, incorporates spatio-temporal features to alleviate data scarcity and improve performance by 5\%.
Additionally, DCGAN uses generated EEG for epilepsy seizure prediction, improving sensitivity of the model \cite{c:18}. 
However, these methods have not been proven to improve the generalization performance of the model in cross data scenarios.


In terms of PD detection, earlier machine learning methods use power spectral density (PSD) or Hjorth features for classification \cite{c:23,c:32}.
Recently, deep neural networks are applied to avoid manually feature extraction and achieve better performance.
A multi-layer CNN structure is developed early \cite{c:11}.
Later, a more deep and advanced networks is proposed, which has 20-layer CNN to distinguish between medicated and non-medicated PD patients \cite{c:22}. 
In addition, ASGCNN \cite{c:24} is designed as a special graph neural network (GNN) with attention modules. 
MCPNet \cite{c:14} is a multi-scale convolutional prototype network to enhance feature diversity and improve model generalization. 
However, these methods are conducted on one single dataset or need pre-training instead of zero-shot cross-dataset settings. 


Different from these existing methods, GEPD not only uses data augmentation to expand EEG datasets for PD but also employs zero-shot learning combined with data distribution analysis to achieve generalizable modeling.

\section{Methods}

\begin{figure}[th]
	\centering
	\includegraphics[width=1\textwidth]{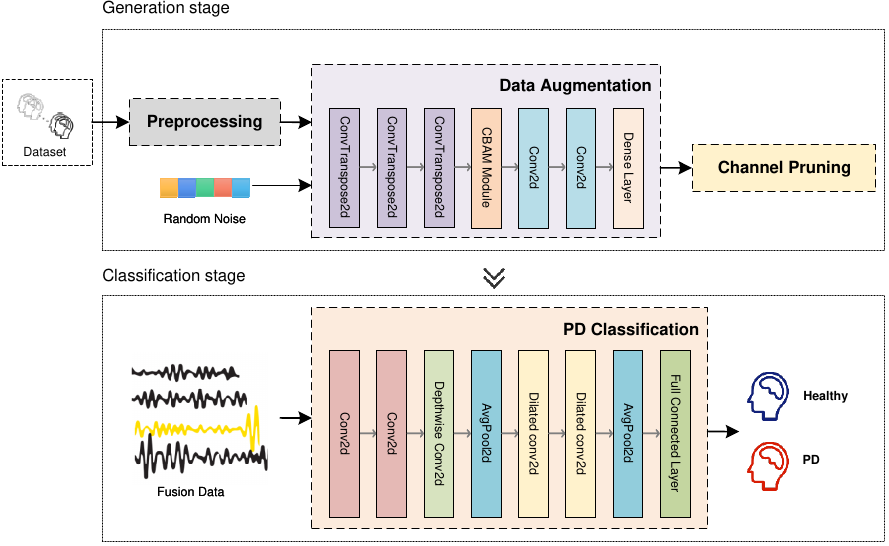} 
	\caption{Overall framework of GEPD: a GAN-enhanced generalizable model for EEG-based detection of Parkinson's disease.
	}
	\label{fig1}
\end{figure}

The overall framework of GEPD is illustrated in Figure \ref{fig1}, which consists of three main components: data augmentation, channel pruning, data quality assessment and PD classification.
First, GEPD uses a designed generative network, called PD-WGAN, to augment EEG data to produce fusion data. 
Then, it optimizes the dataset using a channel pruning method to widen the distribution gap between diseased and healthy data.
In addition, we designed an autoencoder to evaluate the quality of the generated data.
Finally, we design a generalizable PDNex model to detect PD.


\subsection{Data Augmentation}

To address the challenge posed by the small size of EEG datasets, a GAN-based network, PD-WGAN, is designed to augment EEG data. It consists of a generator and a discriminator.
The generator includes one linear layer, three transposed convolutional layers, and one convolutional block attention module (CBAM) layer. 
The linear layer and the transposed convolutional layers are used to generate EEG signals from noise, while the CBAM layer is applied to enhance the features of the generated signals.
The generated data from the generator are passed to the discriminator for evaluation. 
The discriminator includes two convolutional layers, one linear layer, and one CBAM layer. 
These layers evaluate the quality of the generated signals by computing the distance between the generated data and the real data.
Denote the distribution of the real data and the generated data as $p$ and $q$, respectively.
We utilize Wasserstein distance to define this distance $W(p, q)$ as follows:
$$
W(p, q) = \inf_{\pi \in \Gamma(p, q)}E_{(x, y) \sim \pi} [\| x - y \|], \eqno(1)
$$
where \( \Gamma(p, q) \) represents the set of all joint distributions \( \pi \) that transform $p$ into $q$.

In this adversarial process, both the generator and the discriminator continuously optimize and improve. 
Thus, as the discriminator converges, the generator minimizes the loss and generates high-quality data, making it difficult for the discriminator to distinguish between the real data and generated data.

Our dataset contains two groups: healthy control group (HC) and Parkinson's disease group (PD).
During the data augmentation process, PD-WGAN is trained to generate both HC and PD data.
The size of the generated data can be set to be $\delta$ times the original data size. 
The value $\delta$ can only be determined after extensive experimentation.
Then, it combines the generated HC and PD data with the real data to create augmented fusion data.

\subsection{Channel Pruning}
\begin{figure}[t] 
	\centering
	\includegraphics[width= 0.65\textwidth]{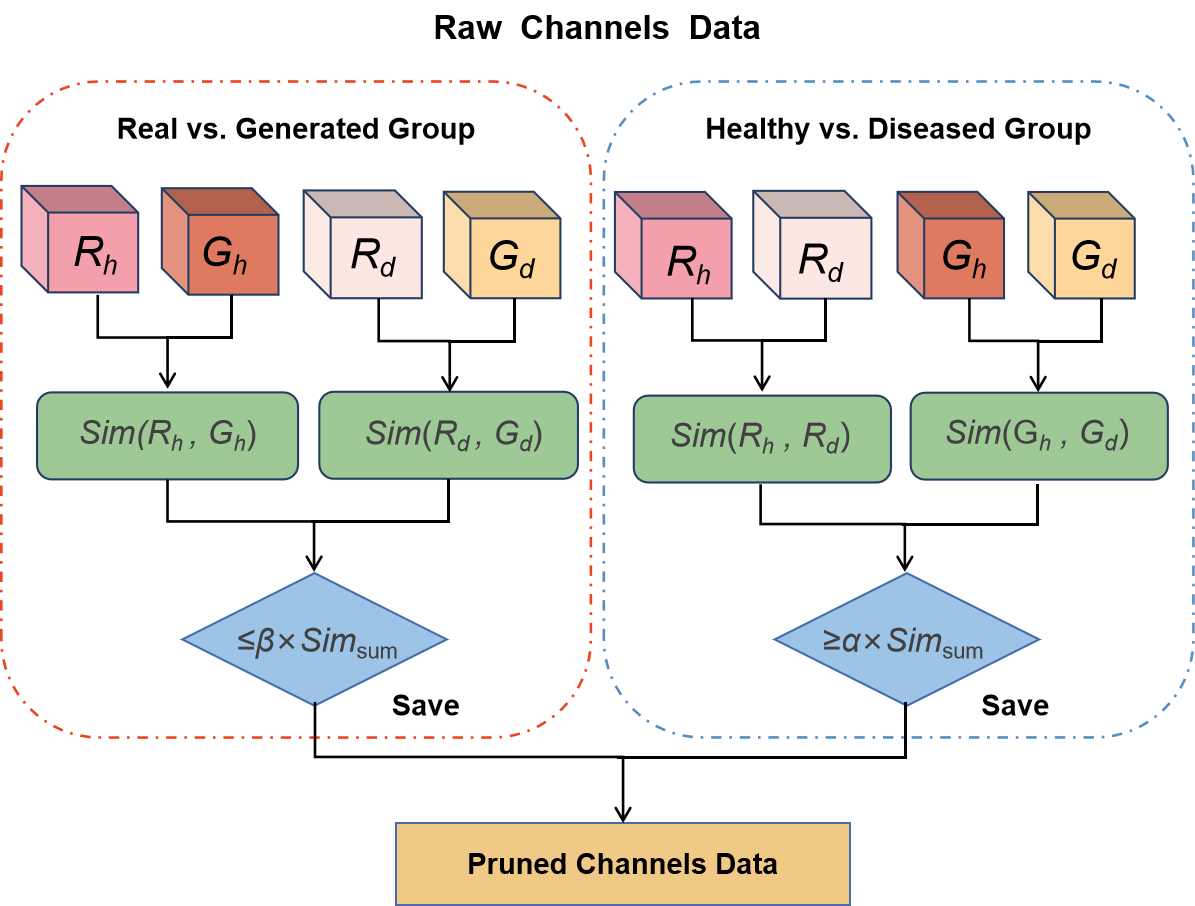} 
	\caption{Schematic diagram of channels pruning.}
	\label{fig2}
\end{figure}

After obtaining the synthetic fusion data using PD-WGAN, we design a channel pruning method to improve the performance of classification, which is shown in Figure \ref{fig2}. 
The channel pruning aims to select the most informative channels data and reduce interfered ones for model training.
First, we divide the fusion data into four categories: 
real healthy control data $R_h$,
real diseased data $R_d$,
generated healthy control data $G_h$, and 
generated diseased data $G_d$.
Then, the pruning process drops the channels that have low similarities between $R_h$ and $G_h$,  and between $R_d$ and $G_d$, which helps to retain characteristic channels that specifically represent EEG patterns of healthy individuals and PD patients, respectively.
In addition, it drops the channels that have high similarities between $R_h$ and $R_d$, and between $G_h$ and $G_d$, which aids in distinguishing between healthy and diseased EEG patterns.

We use Jensen-Shannon $JS$ divergence to formulate the similarity, which is an effective measure of symmetric divergence between two probability distributions.
Jensen-Shannon divergence $JS$ is a smoothed version of Kullback-Leibler $KL$ divergence.
Thus, for two probabilities of $R_h$ and $G_h$, denoted as $P_{R_h}$ and $P_{G_h}$.
The formula of the similarity between $R_h$ and $G_h$, $Sim(R_h, G_h)$, is the $JS$ divergence between $P_{R_h}$ and $P_{G_h}$, which is defined as the average of KL divergence between each distribution:
$$
Sim(R_h, G_h)=JS(P_{R_h}||P_{G_h}), \eqno(2)$$
$$
JS(P_{R_h}||P_{G_h})=\frac{1}{2} KL(P_{R_h}|| M) + \frac{1}{2} KL(P_{G_h}|| M), \eqno(3)
$$
$$
KL(P_{R_h}||P_{G_h}) = \sum_{i} P_{R_h}(i) \log \frac{P_{R_h}(i)}{P_{G_h}(i)}, \eqno(4)
$$
where $M$ is the average of $P_{R_h}$ and $P_{G_h}$.
The calculation of the similarity between other data groups is also referred to Equation (2)-(4).

Next, we set two thresholds $\alpha{Sim}_{sum}$ and $\beta{Sim}_{sum}$ to distinguish between high similarity and low similarity,
where $\alpha$ and $\beta$ are coefficients, ${Sim}_{sum}$ is the sum of JS divergence of the measured data.
Finally, if the similarity between $R_h$ and $G_h$, and between $R_d$ and $G_d$ are smaller than $\beta{Sim}_{sum}$, the corresponding channels will be retained and the remaining channels will be dropped;
if the similarities between $R_h$ and $R_d$, and between $G_h$ and $G_d$ are larger than $\alpha{Sim}_{sum}$, the corresponding channels will be retained and the remaining channels will be dropped.

Therefore, the method of channel pruning removes the channels with insignificant differences between HC and PD groups as well as the channels with excessive differences in the real data and the generated data. The result is a set of channels data with stronger feature attributes that will be used for subsequent classification.

\subsection{PD Classification}
After constructing the fusion data and performing channels pruning, we design a PD classification network, called PDNex, which includes multiple convolutional layers, depthwise convolutional layers, dilated convolutional layers and fully connected layers to extract temporal and spatial features from EEG signals.

Firstly, primary features are extracted through two convolutional layers. 
The first convolutional layer has a kernel of size $1\times32$ with 4 output channels, followed by batch normalization (BN) and ELU activation. 
The second convolutional layer also has a kernel of size $1\times32$ but increases the output channels to 8, similarly followed by BN and ELU activation. 
In an EEG dataset, we denote that it has $N$ channels, $M$ samples, and its batch size is $K$.
Secondly, a depthwise separable convolutional layer is used to further extract global spatial features with a kernel  of size $N\times1$ with 8 groups, and 16 output channels, followed by BN and ELU activation.
Thus, for the \(m\)-th sample data from the \(n\)-th channel in the \(k\)-th batch, 
The depthwise convolution produces an output feature map $O$ of size $M \times N \times K$:
$$O_{m, n, k}=\sum_{i,j}U_{m + i, n + j, k}\cdot \tilde{V}_{i,j,k}, \eqno(5)$$
where $O_{m, n, k}$ is the output of the depthwise convolution at position \((m, n)\) in the \(k\)-th batch,
\( U_{m + i, n + j, k} \) represents the input feature map value at position \((m+i, n+j)\) in the \(k\)-th batch.
\( \tilde{V}_{i,j,k} \) denotes the depthwise convolution kernel of size $i \times j$ for the \(k\)-th batch at position \((m, n)\).


To reduce dimensionality, the model uses an average pooling layer with a pool of size $1\times4$ and introduces a dropout layer to prevent overfitting. 
In the subsequent feature extraction process, the model employs two more depthwise separable convolutional layers. 
These layers have kernels of size $1\times8$ with output channels of 8 and 2, and dilation rates of $1\times2$ and $1\times4$, respectively. The dilated convolution is computed as follows:
\begin{equation}
	\begin{aligned}
		Z_{m,n,k}=\sum_{i,j}V_{m+r\cdot i, n+r \cdot j, k} \cdot \hat{V}_{i,j,k},
	\end{aligned}
	\tag{6}
\end{equation}
where $Z_{m,n,k}$ is the output at position \((m, n)\) in the \(k\)-th batch,
\( V_{m+r\cdot i, n+r \cdot j, k} \) is the input value at position \((m + r \cdot i, n + r \cdot j)\) in the \(k\)-th batch,
\( \hat{V}_{i,j,k} \) is the kernel of size $i \times j$ at position \((m, n)\) for the \(k\)-th batch,
and $r=0.6$ is the dilation rate.
Each feature extraction step includes BN, ELU activation, average pooling (with a pool of size $1\times4$), and dropout layers.

Finally, the multi-dimensional features are flattened into a 1-D vector and mapped to the classification output layer through a fully connected layer. 

\section{Experiments}

\subsection{Datasets}
We use two publicly datasets, UI dataset \cite{ui} and UNM dataset \cite{unm}.
UI includes 28 participants (14 PD patients and 14 healthy controls), and UNM includes 54 participants (27 PD patients and 27 healthy controls). 
All participants are off levodopa medication for at least 12 hours before data collection, and the data are collected in a resting state with eyes open.
It can be seen that the testers come from two different regions, so the two datasets have significant differences, which can be used to verify the generalization performance of our method.

In data preprocessing, the time series of the raw EEG data comes from all non-reference electrodes. 
To unify the reference scheme, we re-reference the UI dataset to the CPz channel according to the international 10–20 system. 
Electrodes not placed in the same locations in both datasets are removed, resulting in the selection of 60 non-reference EEG channels. 
Data are band-pass filtered between 1-45 Hz using a Hamming window band-pass filter. 
Each recording session is divided into uniform, non-overlapping 5-second epochs (2,500 data points per channel), and then normalized using z-score normalization.
These preprocessing steps ensure data consistency and usability, providing a reliable foundation for subsequent analysis and modeling.

\subsection{Settings}
In the generation stage, we use 90\% of the UI dataset for training the PD-WGAN model and the remaining 10\% for testing.
During the adversarial training phase, we conduct 2,000 training epochs with a generator learning rate of 1e-3 and a discriminator learning rate of 1e-4, using Adam optimizer.
The batch size is set to 8. We manually pause training upon discriminator loss convergence and select the model with the minimum generator loss as the final model.
During channel pruning, we set parameters \(\alpha = 0.5\) and \(\beta = 0.5\), reducing the input channels from 60 to 38 for the classification model.

In the classification stage, the batch size for reading data is 8, with input dimensions $8\times38\times2,500$.
Training involves 100 epochs with an initial learning rate of 1e-3, reduced to 1e-4 using cosine annealing.
To prevent overfitting, we apply L1 and L2 regularization coefficients of 1e-4, in conjunction with a dropout rate of 0.6.
Kaiming initialization is used for setting initial weights.
These parameters are based on multiple experiments to achieve the optimal performance. 

\subsection{Data quality assessment}
\begin{table}[t]
	\centering
	\renewcommand{\arraystretch}{1.5}
	\setlength{\tabcolsep}{10pt}
	\begin{tabular}{@{}cccc@{}}
		\toprule
		\textbf{Layer} & \textbf{Type} & \textbf{Input Size} & \textbf{Output Size} \\ \midrule
		Input & - & $(\text{Times}, \text{Channels})$ & $(\text{Times}, \text{Channels})$ \\
		Encoder & Bi-LSTM & $(\text{Times}, \text{Channels})$ & $(\text{Times}, \text{hidden\_size} \times 2)$ \\
		Decoder & LSTM & $(\text{Times}, \text{hidden\_size} \times 2)$ & $(\text{Times}, \text{hidden\_size})$ \\
		Output & Linear (FC) & $(\text{Times}, \text{hidden\_size})$ & $(\text{Times}, \text{Channels})$ \\ \bottomrule
	\end{tabular}
	\caption{Architecture of the Autoencoder}
	\label{table:autoencoder}
\end{table}

\begin{figure}[t]
	\centering
	\includegraphics[width=0.9\textwidth]{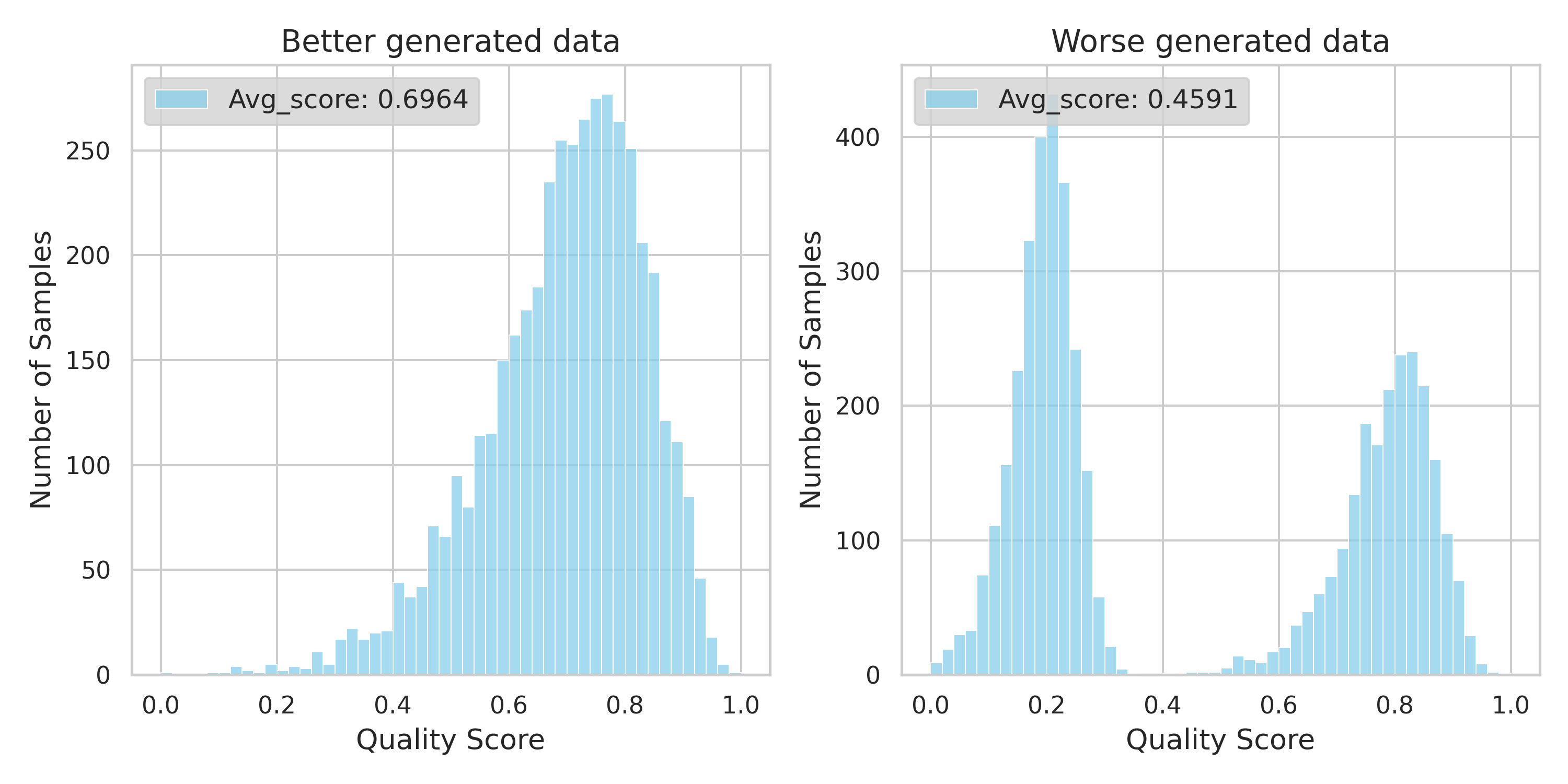} 
	\caption{Histogram of scores for different data quality.}
	\label{fig6}
\end{figure}
Unlike images or text data, the quality of EEG signals is difficult to identify through visual visualization.
In addition, evaluation metrics in the generative domain are inadequate for assessing EEG signal quality.
Thus, we propose an autoencoder-based model to evaluate the quality of generated EEG data using reconstruction error as a measure.
The proposed autoencoder consists of an encoder and a decoder.
The model construction is shown in Table \ref{table:autoencoder}.
The encoder employs a bidirectional LSTM network to capture temporal dependencies in the EEG signals, compressing them into a latent feature vector.
The decoder, a unidirectional LSTM, reconstructs the original signal from this latent representation.
The training process aims to minimize the SmoothL1Loss between the reconstructed signal and the original signal.
For evaluating the quality of generated signals, the generated data is fed into the trained autoencoder, and the reconstruction losses are averaged and normalized. 
Furthermore, we convert the values of losses to a score, which can be  visualized through a histogram. 
A higher score indicates better data quality, allowing an intuitive assessment of the quality of the generated dataset.

Based on the above measure, a representative comparison between high-quality and low-quality data is shown in Figure \ref{fig6}.
The visualized data on the left indicates that better quality data distribution approximates a normal distribution, with an average score greater than 0.65.
In contrast, the data on the right exhibits no discernible pattern, with an average score below 0.5.



\begin{table}[t]
	\centering
	\renewcommand{\arraystretch}{1.7} 
	\setlength{\tabcolsep}{5pt} 
	\resizebox{\columnwidth}{!}{
		\large
		\begin{tabular}{p{4cm}ccccc} 
			\toprule
			\textbf{Models} & \makecell{\textbf{Single-}\\\textbf{dataset}} & \makecell{\textbf{Cross-}\\\textbf{dataset}} & \makecell{\textbf{Base}\\\textbf{Network}} & \makecell{\textbf{w/o}\\\textbf{Fusion Data}} & \makecell{\textbf{w/o}\\\textbf{Pruning}} \\
			\midrule
			\textbf{Baseline}  & 0.831 / 0.825 & 0.803 / 0.814 & - / - & - / - & - / - \\
			\textbf{EEGResNet}  & 0.967 / 0.963 & 0.713 / 0.726 & 0.621 / 0.606 & 0.686 / 0.698 & 0.671 / 0.687 \\
			\textbf{EEGNet}  & 0.971 / 0.965 & 0.724 / 0.741 & 0.631 / 0.607 & 0.706 / 0.718 & 0.667 / 0.631 \\
			\textbf{EEG-Inception}  & 0.981 / 0.975 & 0.726 / 0.725 & 0.653 / 0.648 & 0.664 / 0.675 & 0.695 / 0.711 \\
			\textbf{EEG-ITNet} & 0.981 / 0.978 & 0.734 / 0.736 & 0.684 / 0.672 & 0.694 / 0.681 & 0.698 / 0.723 \\
			\textbf{EEGNex} & \textbf{0.984} / 0.981 & 0.753 / 0.745 & 0.701 / 0.712 & 0.734 / 0.742 & 0.721 / 0.742 \\
			\textbf{GEPD(ours)}  & \underline{0.981} / \textbf{0.983} & \textbf{0.843} / \textbf{0.840} & \underline{0.791} / \underline{0.807} & \textbf{0.814} / \textbf{0.816} & \textbf{0.823} / \textbf{0.821} \\
			\bottomrule
		\end{tabular}
	}
	\caption{Comprehensive comparison of six models and ablation study. Each cell presents "Acc / F1".}
	\label{tab:compbaseline}
\end{table}

\subsection{Comparison with Baselines} 
To evaluate the performance of GEPD, we conduct experiments to compare its performance with other models.
We select six models including a baseline \cite{c:13}, EEGResNet \cite{resnet}, EEGNet \cite{Lawhern_2018}, EEG-Inception \cite{eegInception}, EEG-ITNet \cite{itnet}, EEGNex \cite{c:26}.
Baseline model is a CNN-based model for PD classification from EEG data.
The others are transfer-capable models in the field of motor imagery classification.


To establish a cross-dataset scenario, the models are all trained on UI and tested on UNM.
As shown in Table \ref{tab:compbaseline}, the first column (single-dataset) of the table presents the model performance on a single dataset, training and testing on single UI.
The second column (cross-dataset) shows the cross-dataset performance.
To establish the cross-dataset scenario, the models are all trained on UI and tested on UNM.
From the analysis of the table, we conclude that compared to other method, our proposed GEPD architecture not only achieves comparable  performance on single-dataset scenario, but also achieves the best performance on cross-dataset scenario, with an accuracy of 84.3\% and a F1 score of 84.0\%.
It improves the accuracy and F1 score by 4\% and  2.6\%, respectively, comparing to the second-best method.

We analyze that the proposed GEPD has sufficient and selective learning on single dataset,
while the other models exhibit overfitting to the single dataset, which leading to less performance on the cross-dataset setting.
Therefore, the comprehensive comparison demonstrates the generalizability of GEPD and reveals that our design enables effective balance between learning on the scenarios of single dataset and zero-shot cross-dataset.


\begin{table}[t]
		\centering
		\renewcommand{\arraystretch}{1.7}
		\setlength{\tabcolsep}{23pt}
		\label{table:1}
		\begin{tabular}{ccc}
			\toprule
			\textbf{Data Group} & \textbf{60 Channels} & \textbf{38 Channels} \\ 
			\midrule
			\multicolumn{3}{c}{\textbf{Healthy vs. Diseased Group}} \\ 
			Real-HC/Real-PD & 0.58($\pm0.02$) & \textbf{0.79}($\pm0.01$) ($\uparrow$) \\ 
			\multicolumn{3}{c}{\textbf{Real vs. Generated Group}} \\ 
			Real-HC/Fake-HC & 0.49($\pm0.02$) & \textbf{0.31}($\pm0.01$) ($\downarrow$) \\ 
			Real-PD/Fake-PD & 0.35($\pm0.02$) & \textbf{0.25}($\pm0.01$) ($\downarrow$) \\ 
			\bottomrule
		\end{tabular}
		\caption{Comparison of JS Divergence with and without channel pruning.}
\end{table}


\subsection{Effect of Fusion Data} 
We evaluate the efficacy of the proposed method of data augmentation.
We mute this method and keep the other settings to conduct all the models.
The training data is divided into the original dataset (UI) and a fusion dataset , which is a mix of the original UI data and generated data by PD-WGAN.
Additionally, we also conduct a base network without both the fusion data and channel pruning method as a reference.

As shown in Table \ref{tab:compbaseline}, 
without the fusion data in data augmentation, PDNex still achieves the best performance with an accuracy of 81.4\% and an F1-score of 81.6\%, compared to that of the baseline models.
In addition, the proposed method of data augmentation with fusion data improves the accuracy from 81.4\% to 84.3\%, representing an increment of 2.9\%, and improves 
the F1-score from 81.6\% to 84.0\%, representing an increment of 2.4\%.
These results support the aim of data augmentation method to generate data from existing dataset and construct fusion data for training in order to improve classification performance.

\begin{figure}[t]
	\centering
	\includegraphics[width=0.8\columnwidth]{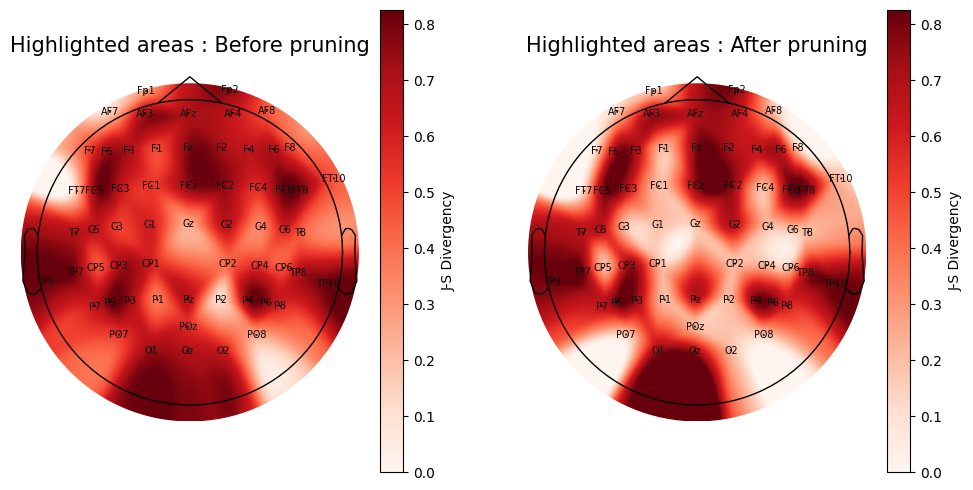} 
	\caption{Visualization of comparison between diseased brain heatmaps before and after channel pruning.}
	\label{fig4}
\end{figure}

\subsection{Effect of Channel Pruning} 
We evaluate the efficacy of the proposed method of channel pruning.
We ablate this method and keep the other settings to conduct all the other models.
We also conduct the base network without both the fusion data and channel pruning method as a reference.
As shown in Table \ref{tab:compbaseline}, 
without the channel pruning method, PDNex has an accuracy of 82.3\% and an F1-score of 82.1\%, which is better than that of the baseline models.
Moreover, the proposed pruning method improves to the accuracy from 82.3\% to 84.3\%, representing an increment of 2.0\%, and similarly improves  
the F1-score from 82.1\% to 84.0\%, representing an increment of 1.9\%.
The results demonstrate the efficacy of channel pruning method in enhancing classification performance.

In Table \ref{table:1}, we show the values of JS divergence-based similarity, comparing the conditions with and without channel pruning.
In PD-WGAN, we train two generative models, respectively, for the EEG data augmentation of HC and PD.
In the result comparison stage, we control two sets of variables.
The first set of variables performs calculating the JS divergence between real HC data and real PD data when using completely real data, with an average value of 0.58 (±0.02).
To maximize channel differences, we remove channels with a JS divergence smaller than this average value, and the recalculated average JS divergence is  0.79 (±0.01).
The second set of variables compare the differences between real data and generate data within the same data group.
The results show that the JS divergence between real HC data and generated HC data is 0.49 (±0.02), and the JS divergence between real PD data and generated PD data is 0.35 (±0.02).
To avoid the generated data becoming useless information, we prune channels with a JS divergence greater than the average value of the second set. This reduces the adjusted JS divergence to 0.31 (±0.01) for HC data and 0.25 (±0.01) for PD data, respectively, and the number of channels is reduced from 60 to 38.

In addition, by visualizing JS divergence based similarity values, we can identify brain regions of interest related to PD, as shown in Figure \ref{fig4}.
These experimental results demonstrate that by controlling and pruning the JS divergence, we can generate high-quality EEG data that are statistically highly similar to real data.
Meanwhile, reducing the number of channels allows us to focus on more representative EEG signals, providing a reliable foundation for EEG data augmentation and analysis using PD-WGAN. This has significant implications for the non-invasive diagnosis and monitoring of PD. 

\begin{figure}[t]
	\centering
	\includegraphics[width=0.75\textwidth]{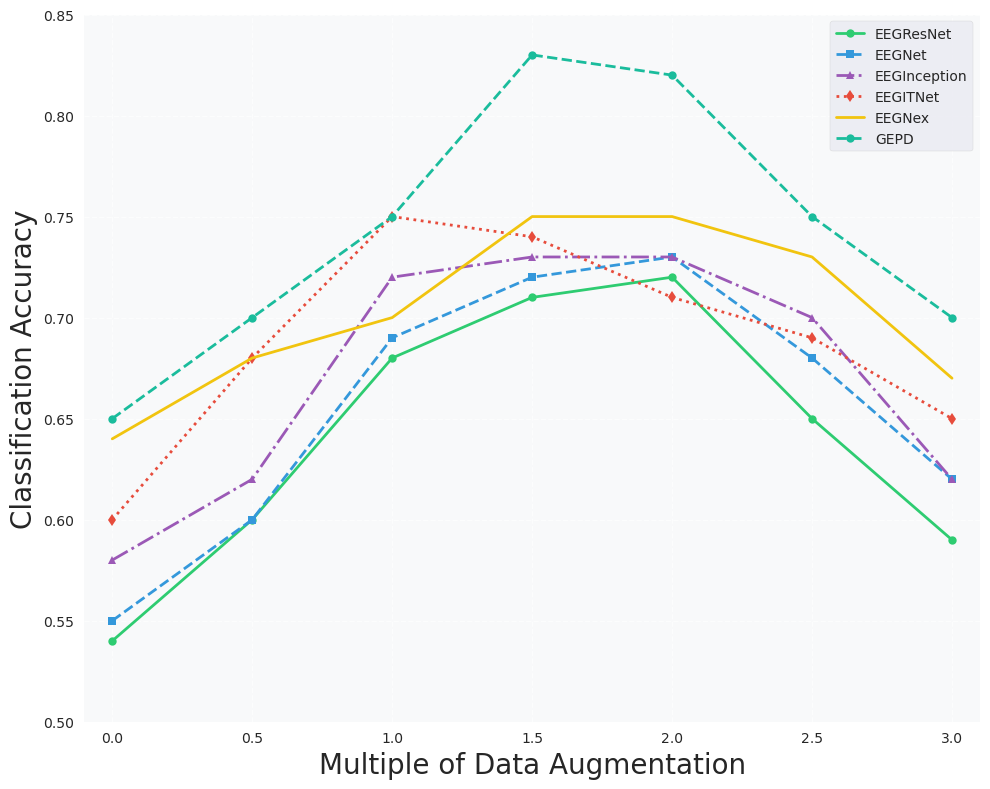} 
	\caption{Performance of different scales of fusion data for PD classification.}
	\label{fig5}
\end{figure}

\subsection{Scale of Data Generation}
In data augmentation, PD-WGAN generates different multiples $\delta$ of more data than the original data,
which results in different scale of fusion data.
To find the best $\delta$ in the classification, 
we vary $\delta$ from 1 to 3 and conduct the experiment with all the baseline models.
As shown in Figure \ref{fig5}, 
expanding the size of the data  from 1 times to 3 times,
we observe that 
different values of $\delta$ result different classification accuracies.
Along with the increment of $\delta$, the trend of accuracy lines are ascending first and then descending.
All the models achieve their best performance when the data is expanded by 1 to 2 times,
and PDNex achieves the best performance when the data is expanded by 1.7 times with an accuracy above 80\%.


%
%

\section{Conclusion and Future Work}
This paper proposes a novel model for PD detection, called GEPD, which leverages a PD-WGAN generative network for data augmentation to address the issue of small EEG dataset sizes and a PDNex network for generalizable classification.
We design a channel pruning method to focus on the most significant EEG channels related to the disease, thereby reducing data complexity while enhancing model performance.
We conduct evaluation experiments and the results demonstrate that GEPD achieves an accuracy of 84.3\% and an F1-score of 84.0\%, which is greater than that of four baseline models.
Furthermore, the results indicate the generalizability of GEPD and reveal its effective balance between scenarios of single dataset and cross-dataset.

In the future work, we plan to explore more methods for EEG data augmentation, such as diffusion models, and enhance interpretability to develop larger model methodologies in the field of EEG disease detection.
Moreover, we plan to explore more quantifiable metrics explaining the pathological causes of neurological diseases.


\end{document}